\documentclass{article}

\usepackage{microtype}
\usepackage{graphicx}
\usepackage{subfigure}
\usepackage{booktabs} 

\usepackage{hyperref}

\usepackage[accepted]{icml2021}

\usepackage{multirow}
\usepackage{amsthm}
\usepackage{mathtools, nccmath}
\usepackage{listings}
\usepackage{color}

\definecolor{codegreen}{rgb}{0,0.6,0}
\definecolor{codegray}{rgb}{0.5,0.5,0.5}
\definecolor{codepurple}{rgb}{0.58,0,0.82}
\definecolor{backcolour}{rgb}{0.95,0.95,0.92}
\lstdefinestyle{pythonstyle}{
    backgroundcolor=\color{backcolour},   
    commentstyle=\color{codegreen},
    keywordstyle=\color{magenta},
    numberstyle=\tiny\color{codegray},
    stringstyle=\color{codepurple},
    basicstyle=\ttfamily\footnotesize,
    breakatwhitespace=false,         
    breaklines=true,                 
    captionpos=b,                    
    keepspaces=true,                 
    numbers=left,                    
    numbersep=5pt,                  
    showspaces=false,                
    showstringspaces=false,
    showtabs=false,                  
    tabsize=2
}
\usepackage{amsmath,amsfonts,bm}

{}
{}
{}
{}

\def\eqref#1{equation~\ref{#1}}

\def\1{\bm{1}}

\def\va{{\bm{a}}}
\def\vb{{\bm{b}}}

\def\ve{{\bm{e}}}

\def\vk{{\bm{k}}}

\def\vq{{\bm{q}}}
\def\vr{{\bm{r}}}

\def\vv{{\bm{v}}}
\def\vw{{\bm{w}}}
\def\vx{{\bm{x}}}
\def\vy{{\bm{y}}}
\def\vz{{\bm{z}}}

\def\mA{{\bm{A}}}

\def\mK{{\bm{K}}}

\def\mR{{\bm{R}}}

\def\mV{{\bm{V}}}
\def\mW{{\bm{W}}}

\DeclareMathAlphabet{\mathsfit}{\encodingdefault}{\sfdefault}{m}{sl}
\SetMathAlphabet{\mathsfit}{bold}{\encodingdefault}{\sfdefault}{bx}{n}

\newcommand{\ReLU}{\mathrm{ReLU}}
\newcommand{\ELU}{\mathrm{ELU}}

\newcommand{\softmax}{\mathrm{softmax}}
\newcommand{\sigmoid}{\sigma}

\theoremstyle{definition}

\icmltitlerunning{Linear Transformers Are Secretly Fast Weight Programmers}

\begin{document}

\twocolumn[
\icmltitle{Linear Transformers Are Secretly Fast Weight Programmers}



\icmlsetsymbol{equal}{*}

\begin{icmlauthorlist}
\icmlauthor{Imanol Schlag$^*$}{idsia}
\icmlauthor{Kazuki Irie$^*$}{idsia}
\icmlauthor{J\"urgen Schmidhuber}{idsia}
\end{icmlauthorlist}

\icmlaffiliation{idsia}{The Swiss AI Lab IDSIA, USI \& SUPSI}

\icmlcorrespondingauthor{Imanol Schlag}{imanol@idsia.ch}
\icmlcorrespondingauthor{Kazuki Irie}{kazuki@idsia.ch}
\icmlcorrespondingauthor{J\"urgen Schmidhuber}{juergen@idsia.ch}

\icmlkeywords{Machine Learning, ICML}

\vskip 0.3in
]



\printAffiliationsAndNotice{\icmlEqualContribution} 

\begin{abstract}
We show the formal equivalence of linearised self-attention mechanisms and fast weight controllers from the early '90s, where a ``slow" neural net learns by gradient descent to program the ``fast weights" of another net through sequences of elementary programming instructions which are additive outer products of self-invented activation patterns (today called keys and values).
Such Fast Weight Programmers (FWPs) learn to manipulate the contents of a finite memory and dynamically interact with it.
We infer a memory capacity limitation of
recent linearised softmax attention variants,
and replace the purely additive outer products by a delta rule-like programming instruction, such that the FWP can more easily learn to correct the current mapping from keys to values. The FWP also learns to compute dynamically changing learning rates.
We also propose a new kernel function to linearise attention
which balances simplicity and effectiveness.
We conduct experiments on synthetic
retrieval problems as well as standard machine translation
and language modelling tasks which demonstrate the benefits of our methods.
\end{abstract}

\section{Introduction}
\label{sec:intro}
Transformers \citep{trafo} have achieved impressive results in a myriad of sequence processing tasks,
including machine translation, language modelling \citep{al2018character, dai2019transformerxlacl, baevski2018adaptive, gpt2}, and question answering \cite{devlin2019bert},
domains previously dominated by recurrent neural networks \citep{graves2013generating, bahdanau2014neural}.

The core component of a Transformer is the self-attention mechanism \citep{cheng16, ParikhT0U16, lin2017structured}
which was recently connected to the modern Hopfield network \citep{ramsauer2021hopfield, KrotovH16, demircigil2017model}.
It extends a form of attention \citep{bahdanau2014neural} 
originally introduced to complement recurrent neural networks, e.g.,~\citep{hochreiter1997long}. 
While relinquishing the recurrence property, all computations across the time axis
can be parallelised.
However, this comes with drawbacks:
self-attention computations scale quadratically with sequence length
while the memory of the model grows linearly.
Therefore, practitioners are forced to limit the context window to a reasonable size,
which in turn makes it impossible to capture longer-term dependencies.

Recent work proposed ``linear Transformers" with constant size memory and time
complexity linear in sequence length  \citep{katharopoulos2020transformers, choromanski2020rethinking, peng2021random, shen2018efficient}.
This complexity reduction is mainly due to 
 a linearisation of the softmax (reviewed in Sec.~\ref{sec:linearAttention}).

Here we emphasize the formal equivalence of this family of linear Transformers and the Fast Weight Controllers or Fast Weight Programmers (FWPs) 
from the '90s \citep{Schmidhuber:91fastweights, schmidhuber1992learning, schmidhuber1993reducing, schmidhuber2021fwp} (apart from normalisation). 
The memories of such FWPs contain key-value associations, and an FWP can learn to reprogram them through sequences of differentiable elementary instructions (also called update rules), which are additive outer products between keys and values invented by the FWP.

This view allows us to derive a limitation of the memory capacity of linear Transformers and similar models.
When the sequence length exceeds storage capacity, 
the model may end up in an overcapacity regime (discussed in depth in Sec.~\ref{sec:capacity}).
To properly operate under such a regime,
the model should learn to dynamically interact with the memory contents and selectively
decide which key-value associations to keep and which ones to delete.
The purely additive instruction may be inappropriate for this purpose.
Therefore, 
inspired by recent work on FWPs
\citep{schlag2020fastweightmemory}, we introduce an improved programming instruction
akin to the famous error-correcting delta-rule \citep{widrow1960adaptive}. 

Furthermore, softmax linearisation techniques for Transformers
are still underexplored.
The existing techniques are either very simplistic \citep{katharopoulos2020transformers}
or mathematically well explained but complex \mbox{\citep{choromanski2020rethinking, peng2021random}}.
We provide a comprehensive comparison and propose
a new method which is both simple and effective.

We demonstrate the benefits of the proposed methods on our own synthetic retrieval dataset (Sec.~\ref{sec:synthetic}), the standard WMT14 English to German machine translation task (Sec.~\ref{sec:translation}), and the Wikitext-103 \citep{merity2016pointer} language modelling task (Sec.~\ref{sec:lm})\footnotemark[2].
\footnotetext[2]{Source code used in this paper is available at \href{https://github.com/ischlag/fast-weight-transformers}{\textit{github.com/ischlag/fast-weight-transformers}}.}

\section{Background on Fast Weight Programmers}
\label{sec:fastweights}
Here we review the concepts of \textit{Fast Weight Programmers} (FWPs)
before relating them to linear Transformer variants in Sec.~\ref{sec:relation_to_transformers}.

In standard neural networks, the weights remain fixed after training,
unlike the activations, which change depending on the inputs at test time.
The general idea of \textit{fast weights} is to make the weights also 
variable and input-dependent.
This concept was called \textit{synaptic modulation} 
\citep{von1981correlation},
a method for variable binding in neural networks
(see e.g.~the recent survey by \citet{greff2020binding}),
or \textit{dynamic connections} \citep{feldman1982dynamic}.
Von der Malsburg defines the effective weights as a (multiplicative) superposition of conventional,
context-independent \textit{slow weights}, and fast changing,
context-dependent \textit{fast weights}.
 \citet{hinton1987using} studied
a net with  (additive) superposition of two sets of weights
with two different learning rates
in a scenario of model retraining. Before 1991, however, no network learned by gradient descent to quickly compute the changes of the fast weight storage of another network or of itself.

Context-dependent FWPs were
introduced in two-network systems of the early '90s \citep{Schmidhuber:91fastweights, schmidhuber1992learning, schmidhuber1993reducing,schmidhuber2021fwp}.
A traditional slow net with slow weights continually changes or reprograms the fast weights of a fast net,
making the fast weights effectively dependent on the spatio-temporal context of a given input stream.
Simply put, the slow net learns to program its fast net. 
Among the proposed elementary differentiable instructions that the slow net can use to program the fast weights,
a particularly attractive one makes use of outer products \citep{Schmidhuber:91fastweights, schmidhuber1992learning}:
for a sequential input $\{ \vx^{(i)} \}_{i=1}^L, \vx^{(i)} \in \mathbb{R}^{d_\text{in}}$,
the model outputs the sequence $\{ \vy^{(i)} \}_{i=1}^L, \vy^{(i)} \in \mathbb{R}^{d_\text{out}}$ as
\begin{eqnarray}
\va^{(i)}, \vb^{(i)} &=& \mW_{a} \vx^{(i)}, \mW_{b} \vx^{(i)} \label{eq:fw1} \\
\mW^{(i)} &=& \sigma\big(\mW^{(i-1)} + \va^{(i)} \otimes \vb^{(i)}\big) \label{eq:fw2} \\
\vy^{(i)} &=& \mW^{(i)} \vx^{(i)} \label{eq:fw3}
\end{eqnarray}
where $\otimes$ denotes the outer product, $\sigma$ is an activation function, $\mW_{a}$ and $\mW_{b}$ are trainable slow weights, while the fast weights $\mW^{(i)}$ are generated at each time step $i$ and serve as a short-term memory.
This is a key-value associative memory model in which the write operation is based on a summation (Eq.~\ref{eq:fw2})
and the retrieval is a matrix-vector multiplication (Eq.~\ref{eq:fw3}).
\citet{schmidhuber1993reducing} describes a recurrent version and discusses ``internal spotlights of attention" (such attention terminology is now widely used in the context of transformers).
The use of outer products results in a model of associations similar to tensor product presentations \cite{smolensky1990tensor}.
In fact, outer-product based associative memory can be found in
numerous works since Hebb's informal rule \citep{hebb1949organization} and its more concrete formal variants  \citep{steinbuch61, steinbuchP63, kohonen72, palm1980associative} including Hopfield networks \citep{hopfield1982neural, little1974existence} and bi-directional associative nets \citep{kosko1988bidirectional}. However, these authors described pre-wired rules to associate given patterns with each other. Their systems did not learn to use such rules for associating self-invented patterns like the FWPs since 1991.

The concept of FWPs has been
revisited recently  \citep{ba2016using, schlag2017gated},
also under different names, e.g.,
 hypernetworks \citep{ha2017hypernetworks, perez2018film, galanti2020hyper},
dynamic plasticity \citep{pmlr-v80-miconi18a, miconi2018backpropamine}, dynamic convolution
\citep{KleinWA15, noh2016image, NIPS2016_8bf1211f}, or lambda networks \citep{bello2021lambdanetworks}
used for applications including meta-learning \citep{munkhdalai2017meta, munkhdalai2018metalearning, MunkhdalaiSWT19, kirsch2020meta}.
FWPs recently also improved memory models
through explicit
mechanisms for facilitating the replacement of deprecated information 
and updating associations \citep{schlag2018learning, schlag2020fastweightmemory}.

\section{Relation to Transformers}
\label{sec:relation_to_transformers}
 \citet{ba2016using} 
have already pointed out a relation between a variant of outer product-based FWPs \citep{schmidhuber1993reducing} and attention \citep{bahdanau2014neural}. \citet{katharopoulos2020transformers} have analysed linearised transformers.
We review these derivations, emphasising the relation between Transformers and the FWPs of the previous section.

\subsection{Self-Attention Without Softmax Is a Fast Weight Programmer}
\label{sec:no_softmax}
A self-attention layer in auto-regressive Transformers \cite{trafo}
maps an input sequence $\{ \vx^{(i)} \}_{i=1}^L, \vx^{(i)} \in \mathbb{R}^{d \times1}$
to an output sequence $\{ \vy^{(i)} \}_{i=1}^L, \vy^{(i)} \in \mathbb{R}^{d_\text{value}\times1}$ as
\begin{eqnarray}
\vk^{(i)}, \vv^{(i)}, \vq^{(i)} &=& \mW_k \vx^{(i)}, \mW_v \vx^{(i)}, \mW_q \vx^{(i)} \label{eq:proj} \\
\mK^{(i)} &=& \big[\mK^{(i-1)}, \vk^{(i)}] \in \mathbb{R}^{d_\text{key} \times i} \\
\mV^{(i)}  &=& \big[\mV^{(i-1)} , \vv^{(i)} ] \in \mathbb{R}^{d_\text{value} \times i} \\
\vy^{(i)} &=& \mV^{(i)} \softmax((\mK^{(i)})^{\top} \vq^{(i)}) \label{eq:trafo_softmax}
\end{eqnarray}
where $[\mA, \va]$ denotes the concatenation of vector $\va$ to matrix $\mA$ along the time dimension,
$\softmax$ is applied along the time dimension,
and $\mW_k$, $\mW_v$, $\mW_q$ are trainable weight matrices.
We omit the scaling by $1/\sqrt{d_\text{key}}$ inside the softmax without loss of generality.

Now if we remove the softmax in Eq.~\ref{eq:trafo_softmax} we obtain:
\begin{eqnarray}
\vy^{(i)} &=& \mV^{(i)} \big((\mK^{(i)})^{\top} \vq^{(i)}\big) = \big(\mV^{(i)} (\mK^{(i)})^{\top}\big) \vq^{(i)} \nonumber \\
&=& \big(\sum_{j=1}^{i} \vv^{(j)} \otimes \vk^{(j)}\big) \vq^{(i)}
\end{eqnarray}
Denoting by $\mW^{(i)}$ the corresponding weight matrix generated from key
and value vectors:
\begin{eqnarray}
\mW^{(i)} = \big(\sum_{j=1}^{i} \vv^{(j)}  \otimes \vk^{(j)} )
\end{eqnarray}
we can rewrite Eqs.~\ref{eq:proj}-\ref{eq:trafo_softmax}
such that they directly relate to Eqs.~\ref{eq:fw1}-\ref{eq:fw3} where the activation function
$\sigma$ is the identity function and without query projection $\mW_q$:
\begin{eqnarray}
\vk^{(i)}, \vv^{(i)}, \vq^{(i)} &=& \mW_k \vx^{(i)}, \mW_v \vx^{(i)}, \mW_q \vx^{(i)} \tag{\ref{eq:proj}} \\
\mW^{(i)}  &=& \mW^{(i-1)} + \vv^{(i)}  \otimes \vk^{(i)} \label{eq:fw_add} \\
\vy^{(i)}  &=& \mW^{(i)} \vq^{(i)} \label{eq:fw_get}
\end{eqnarray}

\subsection{Linearising Self-Attention}
\label{sec:linearAttention}
Instead of removing the softmax as in Sec.~\ref{sec:no_softmax},
prior works have introduced techniques for linearising the softmax \citep{tsai2019},
which has been shown to improve computational efficiency of self-attention for long sequences \citep{katharopoulos2020transformers, choromanski2020rethinking, peng2021random}.

By writing the softmax explicitly, Eq.~\ref{eq:trafo_softmax} can be written as:
\begin{align}
    \label{eq:softmax}
    \vy^{(i)} = \sum_{j=1}^i \frac{
         \vv^{(j)}  \kappa(\vk^{(j)}, \vq^{(i)})
         }{
         \sum_{j'=1}^i \kappa(\vk^{(j')}, \vq^{(i)})
         }
\end{align}
where $\kappa(\vk, \vq) = \exp(\vk \cdot \vq) \in \mathbb{R}_{>0}$ is the softmax kernel
and $\vk \cdot \vq = \vk^{\top}\vq$ is the vector dot product.

The general idea is to replace the softmax kernel $\kappa$ by another kernel:
$\kappa'(\vk, \vq) = \phi(\vk)^{\top} \phi(\vq)$
where $\phi$ is a function $\mathbb{R}^{d_\text{key}} \rightarrow \mathbb{R}^{d_\text{dot}}$.
We discuss the necessary properties of $\phi$ in Sec.~\ref{sec:phi_prop}.
By replacing $\kappa$ in Eq.~\ref{eq:softmax} by $\kappa'$, we obtain
\begin{align}
\label{eq:linearSM}
    \vy^{(i)} & = \sum_{j=1}^i  \frac{
         \vv^{(j)} \phi(\vk^{(j)})^{\top} \phi(\vq^{(i)})
         }{
         \sum_{j'=1}^i \phi(\vk^{(j')}) \cdot \phi(\vq^{(i)})
         } \\
    & = \displaystyle \frac{ \sum_{j=1}^i
         \big(\vv^{(j)} \phi(\vk^{(j)})^{\top} \big) \phi(\vq^{(i)})
         }{
         \big(\sum_{j'=1}^i \phi(\vk^{(j')})\big) \cdot \phi(\vq^{(i)})
         }
\end{align}
Using the outer-product notation,
the numerator is analogous to the case without softmax (Sec.~\ref{sec:no_softmax}):
\begin{eqnarray}
 \displaystyle \sum_{j=1}^i
         \big(\vv^{(j)} \phi(\vk^{(j)})^{\top} \big) \phi(\vq^{(i)}) = \displaystyle \big(\sum_{j=1}^i
         \vv^{(j)} \otimes \phi(\vk^{(j)}) \big) \phi(\vq^{(i)}) \nonumber
\end{eqnarray}
By introducing the fast weight matrix $\mW^{(i)}$ and an additional vector $\vz^{(i)}$ for the denominator,
\begin{eqnarray}
\mW^{(i)} &=& \sum_{j=1}^i \vv^{(j)} \otimes \phi(\vk^{(j)}) \\
\vz^{(i)} &=& \displaystyle \sum_{j=1}^i \phi(\vk^{(j)})
\end{eqnarray}
forward computations of linear Transformers can be written as \citep{katharopoulos2020transformers}:
\begin{eqnarray}
\vk^{(i)}, \vv^{(i)}, \vq^{(i)} &=& \mW_k \vx^{(i)}, \mW_v \vx^{(i)}, \mW_q \vx^{(i)}  \tag{\ref{eq:proj}}  \\
\mW^{(i)}  &=& \mW^{(i-1)} + \vv^{(i)}  \otimes \phi(\vk^{(i)}) \label{eq:fw_phi_add} \\
\vz^{(i)}  &=& \vz^{(i-1)}+ \phi(\vk^{(i)}) \\
\vy^{(i)}  &=& \displaystyle \dfrac{1}{\vz^{(i)} \cdot \phi(\vq^{(i)})} \mW^{(i)} \phi(\vq^{(i)}) \label{eq:fw_phi_get}
\end{eqnarray}
which is a Fast Weight Programmer (Sec.~\ref{sec:fastweights}) with normalisation.
Hence, the core of  linear Transformer variants 
are outer product-based Fast Weight Programmers.

\section{Analysing and Improving Linear Transformers as Fast Weight Programmers}
Viewing linear Transformer variants as Fast Weight Programmers provides us with two insights which we investigate in this work:
their capacity limits as associative memories (Sec.~\ref{sec:capacity}), 
and their ineptness to edit previously stored associations (Sec.~\ref{sec:updaterule}).

\subsection{Capacity Limitation}
\label{sec:capacity}
\paragraph{Intuition.}
Endlessly adding new associations to a memory of finite size, as in Eq.~\ref{eq:fw_phi_add}, inevitably will reach a limit. 
In linear attention, information is stored in a matrix and is retrieved using matrix multiplication (see Eq.~\ref{eq:fw_phi_get}).
As a consequence, to prevent associations from interfering with each other upon retrieval, the respective keys need to be orthogonal. 
Otherwise, the dot product will attend to more than one key and return a linear combination of values.
With keys embedded in a $d_\text{dot}$ space, there cannot be more than $d_\text{dot}$ orthogonal vectors.
That is, storing more than $d_\text{dot}$ associations will result in a retrieval error. 
In linear Transformers, when the length of the sequence is longer than $d_\text{dot}$,
the model might be in such an overcapacity regime.
While we experimentally demonstrate this effect on toy tasks (Sec.~\ref{sec:synthetic}), prior work on tensor product representations allows for a more formal discussion.

\paragraph{Tensor Product Representation Theory.}
Early work in connectionist research investigated the usage of distributed representations as a means for storing symbolic structures. 
One highly-influential work is the tensor-product-based variable binding mechanism \citep{smolensky1990tensor}.
A tensor product representation (TPR) of a structured symbolic system consisting
of a set of variables and values constructed
from outer products of the so called \textit{role} and \textit{filler} vectors.
These terms directly translate into \textit{keys} and \textit{values} in our context.
The fast weight memories of Eq.~\ref{eq:fw_phi_add} are the most
basic form of such representations (second order tensors).
Therefore, many results discussed in Smolensky's work transfer to our model.
In particular, Theorem 3.3 and 3.1 of \citet{smolensky1990tensor} discuss more formally the \textit{crosstalk} and retrieval error intuitively described in the previous paragraph.

However, we also note an important difference:
the classic TPRs of \citet{smolensky1990tensor}
are constructed with a priori knowledge of the symbolic structure.
In contrast, our FWPs since 1991, including  recent FWPs \citep{schlag2018learning}, learn all the vectors involved in constructing such a representation.

\subsection{Improving the FWP's Programming Instruction}
\label{sec:updaterule}
Sec.~\ref{sec:capacity} argues that the linear Transformers
can end up in an overcapacity regime, if the sequence length $L$ exceeds the
dimension $d_\text{dot}$ of the keys.
Once in overcapacity, an ideal memory model should dynamically interact with
the memory contents and selectively determine which associations to remember or to forget.
This is in stark contrast to the standard Transformer which stores \textit{immutable}
pairs of key and value vectors by concatenation, thus increasing the storage size.
While such models work well in practice, we consider a model's capability to update previously acquired knowledge to be critical for many problems.
Hence, from the perspective of dynamic interaction with the memory, the purely additive update rule of Eqs.~\ref{eq:fw_phi_add}
may be sub-optimal.
This motivates us to improve the elementary differentiable programming instruction (i.e.~the update rule) of FWPs. 

Inspired by the recent work by \citet{schlag2020fastweightmemory}, we propose a basic instruction that essentially implements the famous error-correcting delta rule \citep{widrow1960adaptive} in an end-to-end differentiable way, such that the FWP can learn to use it wisely, through  self-invented, dynamically changing learning rates. 
Given a new input key-value pair $(\vk^{(i)}, \vv^{(i)})$,
the FWP first accesses the current state of the memory $\mW^{(i-1)}$ and retrieves the value $\bar{\vv}^{(i)}$ currently paired with the key $\vk^{(i)}$. Then the model stores
a convex combination $\vv^{(i)}_\text{new}$ of the retrieved value $\bar{\vv}^{(i)}$ and the input $\vv^{(i)}$
using an interpolation weight $0 \leq \beta^{(i)} \leq 1$ also generated by the model.
The model thus sequentially transforms an input sequence $\{ \vx^{(i)} \}_{i=1}^L, \vx^{(i)} \in \mathbb{R}^{d\times1}$
into an output sequence $\{ \vy^{(i)} \}_{i=1}^L, \vy^{(i)} \in \mathbb{R}^{d_\text{value}\times1}$ as:
\begin{align}
\vk^{(i)}, \vv^{(i)}, \vq^{(i)} &= \mW_k \vx^{(i)}, \mW_v \vx^{(i)}, \mW_q \vx^{(i)} \tag{\ref{eq:proj}} \\
\bar{\vv}^{(i)} &= \mW^{(i-1)} \phi(\vk^{(i)}) \label{eq:v_old}\\
\beta^{(i)} &= \sigmoid(\mW_\beta \vx^{(i)}) \\
\vv^{(i)}_\text{new} &= \beta^{(i)} \vv^{(i)} +(1-\beta^{(i)}) \bar{\vv}^{(i)}  \label{eq:v_new}
\end{align}

where $\mW_\beta \in \mathbb{R}^{1 \times d}$, and $\sigmoid$ is the sigmoid function. 
The interpolation weight $\beta^{(i)}$  is the ``write-strength'' as it defines to which extent the new value will replace the previous value.
We note that while $\beta^{(i)}$ only depends on $\vx^{(i)}$,
in a multi-layer model, $\vx^{(i)}$ has the full context information except in
the first layer.
We set $\mW^{(0)}=0$ and $\vz^{(0)}=0$.
Then the fast weight update rule and the final output $\vy^{(i)}$ are defined as follows (see Appendix \ref{app:update_rule} for detailed derivations):
\begin{align}
\label{eq:updaterule}
\mW^{(i)} &= \mW^{(i-1)}
    \underbrace{+ \vv^{(i)}_\text{new} \otimes \phi(\vk^{(i)})}_{\text{write}}
    \underbrace{- \bar{\vv}^{(i)} \otimes \phi(\vk^{(i)})}_{\text{remove}} \\
&= \mW^{(i-1)} + \beta^{(i)}(\vv^{(i)} - \bar{\vv}^{(i)}) \otimes \phi(\vk^{(i)})  \label{eq:updaterule2}
\end{align}
\begin{align}
\vy^{(i)}  &= \mW^{(i)} \phi(\vq^{(i)}) \label{eq:update_out}
\end{align}
As shown in Eq.~\ref{eq:updaterule2}, our programming
instruction or update rule is effectively a delta rule
with a dynamic learning rate $\beta^{(i)}$.
The model thus learns to correct the current
key to value association.
In Appendix \ref{sec:peng}, we formally show
the advantage of this approach over the 
gated update rule concurrently proposed by \citet{peng2021random}.

\paragraph{Normalisation.}
In the equations above, no normalisation is applied to the value we retrieve.
A straightforward normalisation can be obtained by following the derivation in Sec.~\ref{sec:linearAttention}, i.e. by introducing an accumulator:
\begin{align}
\vz^{(i)} &= \vz^{(i-1)} + \phi(\vk^{(i)}) \label{eq:acc}
\end{align}
and replacing Eqs.~\ref{eq:v_old} and \ref{eq:update_out} respectively by:
\begin{align}
\bar{\vv}^{(i)} &= \frac{\mW^{(i-1)} \phi(\vk^{(i)})}{\vz^{(i-1)} \cdot \phi(\vk^{(i)})} \\
\vy^{(i)}  &= \displaystyle \dfrac{\mW^{(i)} \phi(\vq^{(i)})}{\vz^{(i)} \cdot \phi(\vq^{(i)})} 
\end{align}
where we define $\bar{\vv}^{(1)}=0$.
In this approach, the output $\vy^{(i)}$ is a weighted average of $\beta^{(j)}(\vv^{(j)} - \bar{\vv}^{(j)})$ for $1\leq j \leq i$.
We refer to this approach as \textit{attention normalisation}.

This approach, however, has drawbacks.
First, the accumulation of positive values in Eq.~\ref{eq:acc} always grows with the number of steps,
and may result in instability.
Second, specifically for our update rule, 
this normalisation is not sufficient to balance the \textit{weights} between write
and remove operations in Eq.~\ref{eq:updaterule}
(see derivations in Appendix \ref{app:norm}).
Here we propose a better approach based on simple normalisation.
We divide the effective key and query vectors $\phi(\vk^{(i)})$ and $\phi(\vq^{(i)})$
by the sum of its components, e.g., for the query:
\begin{align}
\phi'(\vq^{(i)}) &= \displaystyle \dfrac{\phi(\vq^{(i)})}{ \displaystyle \sum_{j=1}^{d_\text{dot}} \phi(\vq^{(i)})_j}
\end{align}
before applying  Eqs.~\ref{eq:v_old}-\ref{eq:update_out}.
A general consequence of this normalisation is  intuitively understood 
by noticing that the output of any matrix-vector operations
(like Eq.~\ref{eq:update_out}) is a
weighted sum of columns of the matrix where weights are the components
of the vector; thus, if the vector components sum up to one,
the operation can be viewed as an attention over the columns of the matrix.
We provide further explanations and precise
implications for our FWP in Appendix \ref{app:norm}.
We refer to this approach as \textit{sum normalisation}.

Since this is a simple substitution of $\phi(\vk^{(i)})$ and $\phi(\vq^{(i)})$ in Eqs.~\ref{eq:v_old}-\ref{eq:update_out},
one might still ask whether additional attention normalisation is needed.
In language modelling experiments (Sec.~\ref{sec:lm}), we show that this is not the case.

\section{Linear Attention Functions}
\label{sec:phi_section}
The central component of softmax linearisation (Sec.~\ref{sec:linearAttention}) is the $\phi$ function
which maps key and query vectors to the space where the dot product is executed: 
$\mathbb{R}^{d_\text{key}} \rightarrow \mathbb{R}^{d_\text{dot}}$. 
We first list desirable properties of such a function,
and review the existing $\phi$ functions from the perspective of fast weight memories. 
Finally, we also propose our own $\phi$ function.
\subsection{Properties}
\label{sec:phi_prop}
For Eq.~\ref{eq:linearSM} to define proper attention weights
between 0 and 1, the codomain of $\phi$ should be \textbf{positive}.
Another property of $\phi$ derives from the discussion of memory
capacity in Sec.~\ref{sec:capacity}.
The dimensionality of its codomain $d_\text{dot}$ defines the model's capacity.
Therefore, by including a transformation which \textbf{projects} the input dimension $d_\text{key}$
to a larger dimension $d_\text{dot}$, the $\phi$ function can potentially increase the upper bound
of the capacity.

\subsection{Katharopoulos' Linear Attention}
\label{sec:katharopoulos}
\citet{katharopoulos2020transformers} propose to use the simple element-wise $\ELU + 1$ function \citep{clevert2015fast}:
\begin{align}
    \phi(x) = \ELU(x) + 1= 
    \begin{cases}
        x+1,& \text{if } x > 0\\
        \exp(x), & \text{if } x \leq 0
    \end{cases}
\end{align}
The choice of $\ELU$ over $\ReLU$ is motivated by non-zero gradients on the negative part.
Importantly, as a simple element-wise function,
this $\phi$ function preserves the dimension of the input key vector ($d_\text{key}=d_\text{dot}$),
without modifying the memory capacity
as discussed in Sec.~\ref{sec:capacity}.

\subsection{FAVOR+}
\label{sec:performer}
In contrast to \citet{katharopoulos2020transformers}'s $\phi$ function 
which merely satisfies positivity (and a good gradient) property,
\citet{choromanski2020rethinking} propose a mathematically rigorous method
to approximate the softmax with random features.
They propose the following $\phi$ function:
\begin{align}
    h(\vx) &= \frac{1}{\sqrt{2}} \exp(-\frac{||\vx||^2}{2}) \\
    \phi(\vx) &= \frac{h(\vx)}{\sqrt{m}}
    \begin{bmatrix}\exp(\mR \vx) \\ \exp(-\mR \vx) \end{bmatrix} \label{eq:performer}
\end{align} 
where the concatenation $\begin{bmatrix}\va \\ \vb \end{bmatrix}$ of two vectors $\va$ and $\vb$ is along the feature dimension, and $\mR \in \mathbb{R}^{m \times d_\textit{key}}$ is a matrix with $m$ random features where each row vector $\vr \in \mathbb{R}^{1 \times d_\textit{key}}$ is drawn from $\mathcal{N}(0,\mathbf{I}_{d_\text{key}})$.
A similar approach is also proposed by \citet{peng2021random}.

With FAVOR+, the dimension of the codomain $d_\text{dot}$ is $2m$ which increases the theoretical capacity of the memory if $2m > d_\text{key}$. 
At the same time, the model's capacity is still limited, and equals the infinite capacity of the softmax memory only when $m$ goes to infinity, which is never achieved in practice.
During training, we redraw these $m$ random vectors for each mini-batch.
During evaluation, we draw a set of $m$ random vectors once, and keep them fixed.
$m$ is the only hyperparameter of FAVOR+ and influences the quality of the softmax approximation. 
\citet{choromanski2020rethinking} suggest to choose $m$ in the order of $d_\textit{key}\log(d_\textit{key})$.
This sampling process is the main drawback of FAVOR+ as it introduces variance into the model's output.

\subsection{Deterministic Parameter-Free Projection (DPFP)}
\label{sec:phi}
The two previous sub-sections highlight the sub-optimality of the existing $\phi$ functions.
Sampling introduces extra complexity to FAVOR+ (Sec.~\ref{sec:performer}),
while the Linear Transformer (Sec.~\ref{sec:katharopoulos}) lacks the ability to project up the dot product dimension.
Here we propose an alternative approach called \textit{deterministic parameter-free projection} (DPFP).
It is deterministic and easy to compute like Linear Transformers
while increasing the dot product dimension
without requiring FAVOR+'s random features.

\begin{figure}[h]
    \begin{center}
        \centerline{\includegraphics[width=0.55\columnwidth]{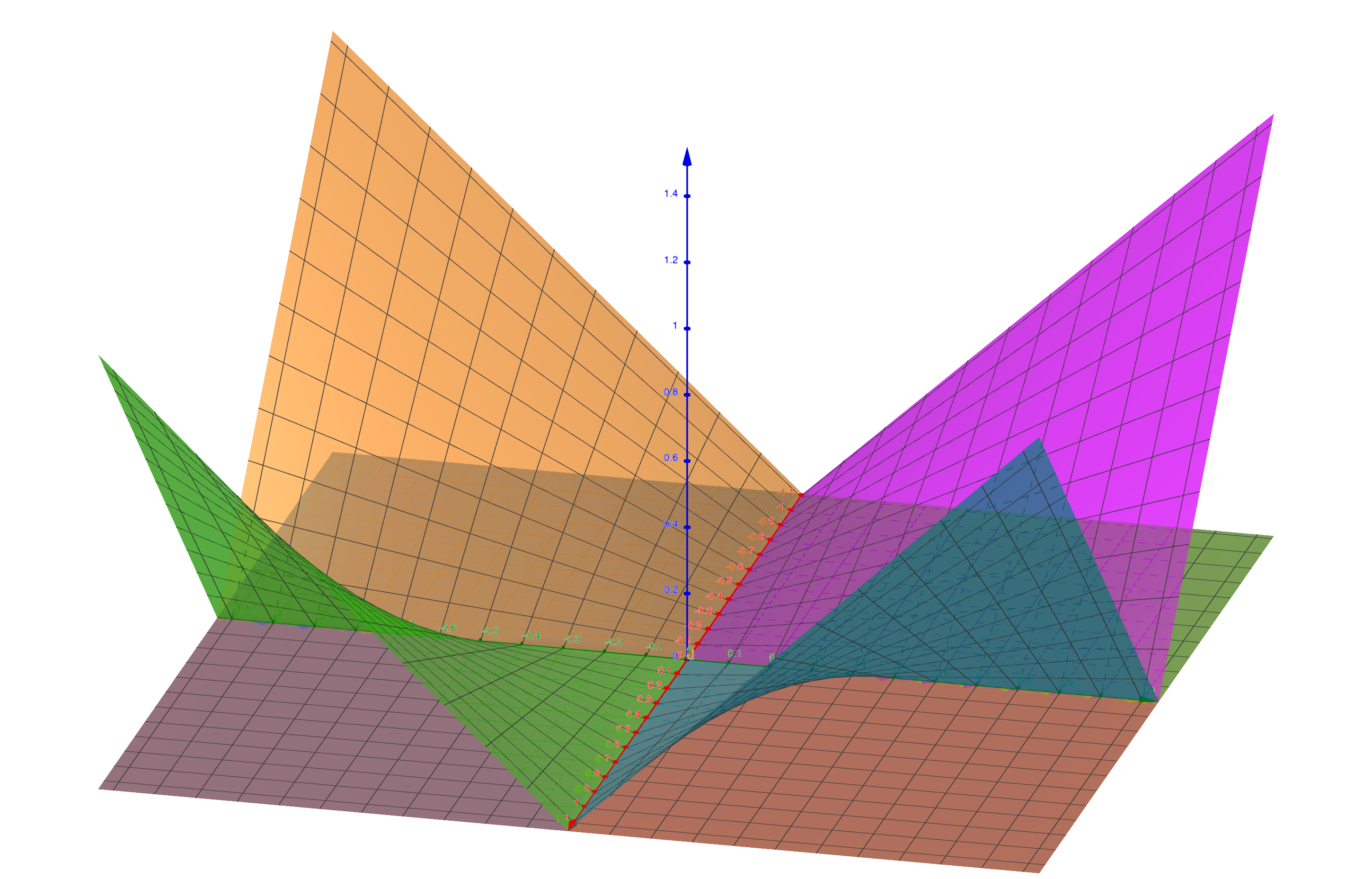}}
        \caption{A visualisation of a DPFP from a 2d space (the xy-plane) to a 4d space (the four colored surfaces). Each surface is a partial function which represents one element of the 4d vector.}
        \label{fig:3dphi}
    \end{center}
    \vskip -0.2in
\end{figure}

We begin with a low-dimensional example to foster an intuitive understanding before moving on to the general formulation. 
Consider 4 keys $\vk^{(i)}, i \in \{1,2,3,4\}$ in $\mathbb{R}^2$ and $\phi: \mathbb{R}^2 \rightarrow \mathbb{R}^4_{\geq0}$ where the $l$-th element of $\phi(\vx)$ is generated by the partial function $\phi_l: \mathbb{R}^2 \rightarrow \mathbb{R}_{\geq0}$. 
We design $\phi$ such that it facilitates orthogonality in the projected space,
i.e. $\phi(\vk^{(i)}) \cdot \phi(\vk^{(j)}) = 0$ for $i \neq j$.
Towards this end, we construct $\phi$ such that if $\phi_l(\vx) > 0$ then $\phi_{n}(\vx) = 0$ for all $n \neq l$.
Such a constraint can be enforced by limiting the domains of the partial functions to be non-overlapping.
With the element-wise rectifier function $r(a) = \max(0,a)$ the partial functions are defined as:
\begin{align}
    \phi_1(\vk) &= r(\vk_1) r(\vk_2) \\
    \phi_2(\vk) &= r(-\vk_1) r(\vk_2) \\
    \phi_3(\vk) &= r(\vk_1) r(-\vk_2) \\
    \phi_4(\vk) &= r(-\vk_1) r(-\vk_2)
\end{align}
Figure \ref{fig:3dphi} illustrates this function.
The elements of the 4-dimensional space are displayed as the $z$ component of the four coloured surfaces.
The figure shows how each vector in the 2d plane will have a single non-zero component in the 4d space and equally splits the input space into four areas which will be orthogonal in the projected space.

We generalise this method to higher dimensional inputs by constructing additional two-factor features.
Given an input vector $\vk \in \mathbb{R}^{d_\text{key}}$ and $i \in [1,2 d_\text{key}]$, the partial function
\begin{align}
    \label{eq:dpfp}
    \phi_{i\nu}(\vk) = 
        r(\begin{bmatrix} \vk \\ -\vk\end{bmatrix})_i 
        r(\begin{bmatrix} \vk \\ -\vk\end{bmatrix})_{i+\nu}
\end{align}
where $\nu \in \{1,2,.., d_\text{key}2 - 1\}$ is a capacity controlling hyperparameter.
The codomain dimensionality of $\phi(\vk)$ is thus $d_\text{dot} = 2d_\text{key}\nu$.
Eq.~\ref{eq:dpfp} is highly parallelisable because each partial function can be computed independently. 
This can be implemented in few lines of code as we show in Appendix \ref{app:dpfp}.

Finally we note that \citet{choromanski2020rethinking} empirically show that replacing $\exp$
in Eq.~\ref{eq:performer} by $\ReLU$ typically improves model performance.
While this result has not been theoretically justified, it supports the
design of our DPFP which aims for sparsity and orthogonality.

\section{Experimental Results}
Now we present our experimental results on
synthetic retrieval problems (Sec.~\ref{sec:setting1} and \ref{sec:setting2}),
 machine translation (Sec.~\ref{sec:translation}), and language modelling (Sec.~\ref{sec:lm}).\looseness-1
\subsection{Synthetic Settings}
\label{sec:synthetic}
We illustrate the capacity issue (Sec.~\ref{sec:capacity}) of 
linear attention and the effectiveness of our new update rule (Sec.~\ref{sec:updaterule}) on two synthetic problems.

In both settings, our toy problem consists of retrieving the correct value from a sequence of randomly sampled key-value associations when queried with one of the used keys.
Crucially, the query is given at the end of the sequence, such that the model is not aware of it
while processing the inputs.
To succeed, the model has to learn to store the observed associations in its memory without interference. 

Let $\mathcal{K}$ and $\mathcal{V}$ be the finite and fixed sets of keys and values and $S=|\mathcal{K}|=|\mathcal{V}|$. 
Then, the input to the model is the sequence 
$[(\mathsf{k},\mathsf{v})_1, ..., (\mathsf{k},\mathsf{v})_L]$ 
followed by $\mathsf{q}$ 
where every pair $(\mathsf{k},\mathsf{v}) \in \mathcal{K} \times \mathcal{V}$ is sampled randomly,
and $\mathsf{q}$ is randomly chosen to be one of the $L$ keys.

Each value $\mathsf{v}^{(i)}, i \in [1,..,S]$ is assigned a fixed one-hot vector $\vv^{(i)} \in \mathbb{R}^S$. Hence, the set of value vectors is an orthonormal basis.
In contrast, the vector embedding of the key symbols is the learned function $e: \mathcal{K} \rightarrow \mathbb{R}^{d_\text{emb}}$
and $\vk = \mW_K [e(\mathsf{k});\vv]$ 
where $\mW_K \in \mathbb{R}^{d_\text{key} \times (d_\text{emb} + S)}$.

Following the $L$ write operations, the read function and the query vector 
$\vq = \mW_Q e(\mathsf{q}), \mW_Q \in \mathbb{R}^{d_\text{key} \times d_\text{emb}}$ 
are used to retrieve $\hat{\vv} \in \mathbb{R}^S$ from memory.
Finally, the loss is defined as 
$l(\hat{\vv},\vv^*) = \sum_j^S \frac{1}{2}(\vv^*_j - \hat{\vv}_j)^2$ 
where $\vv^*$ is the value vector assigned to $\mathsf{q}$ in the input sequence. 
Each model is trained in mini-batches using this loss and Adam with default hyperparameters unless stated otherwise.
For evaluation, we sample 20 sequences and test all possible queries,
e.g., with $S=100$ unique keys, the evaluation batch is of size $100*20=2000$.

\subsubsection{Setting 1: Testing Capacity}
\label{sec:setting1}
In this setting, we experimentally demonstrate the capacity limit of linear attention (Sec.~\ref{sec:capacity}).
We conduct experiments for the various $\phi$ functions described in Sec.~\ref{sec:phi_section}.
We fix $d_\text{key}$ to be $64$, while different $\phi$ functions produce different $d_\text{dot}$.
We set the sequence length to be equal to the number of unique keys ($L = S$),
and sample the keys and values without replacement to generate the sequences.
By varying the sequence length $S$, our goal is to show that all linear attention models 
(using the simple sum update rule of Sec.~\ref{sec:linearAttention}) fail at retrieving when $S$ exceeds $d_\text{dot}$.

\begin{figure}[h]
    \begin{center}
        \centerline{\includegraphics[width=\columnwidth]{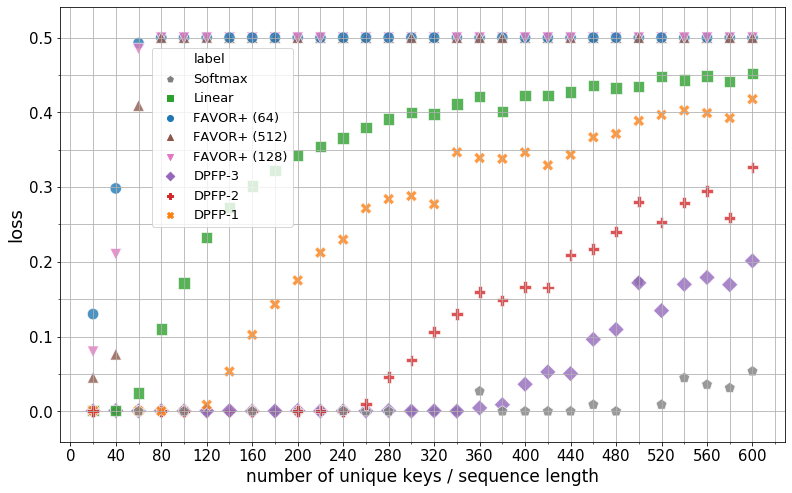}}
        \caption{Final evaluation loss of the softmax memory and various linear attention mechanisms on associative retrieval problems with the total number of unique associations ranging from 20 to 600. Each individual symbol is a model trained until convergence.}
        \label{fig:setting1}
    \end{center}
    \vskip -0.7cm
\end{figure}

All models are trained with a mini-batch size of $32$ until the evaluation loss falls below $0.001$ or until lack of progress for $1000$ steps.
In Figure \ref{fig:setting1}, the best validation set performance for each model and each $S$ is displayed
(for the learning curves see Appendix \ref{app:setting1}).
The number of unique keys is initially $S=20$ and is incremented by $20$ until $S=600$.
The following models are compared: Softmax, Linear-Attention, FAVOR+ with 64, 128, and 512 random features, DPFP-$\nu$ with $\nu \in \{1,2,3\}$.

The results support our theoretical analysis. 
Linear-Attention has a capacity of $64$ due to the choice of $d_\text{key}=d_\text{dot}=64$. 
Experimentally, Linear-Attention begins to accumulate errors with $60$ or more associations. 
Similarly, DPFP projections 1, 2 and 3 start to accumulate errors as they approach their respective limits at $128$, $256$, and $384$.
FAVOR+, on the other hand, fails to achieve a loss of 0 in any experiment.
Finally, as expected, softmax attention is outperforming all $\phi$ functions, although it struggles to fully converge with more than 500 keys.

\subsubsection{Setting 2: Comparing Update Rules}
\label{sec:setting2}
In the second setting, we compare variations of the update rule.
Unlike in setting 1, keys and values will be sampled with replacement and sequence length $L = 2S$. 
As a result, in the same sequence, multiple keys can be re-assigned to a new value more than once.
The expected value to retrieve is the most recent one associated with the query.
With every new key, the previous value associated with this key deprecates and the model is required to update its finite size memory. 
The ability to update values associated with keys is essential 
to bind context-specific values to a key.

We use DPFP-1 as the $\phi$ function.
The sequence length is fixed at 40 with 20 unique keys and values.
While this setting does not exceed the capacity of DPFP-1, our result
is independent of the capacity regime
(see results for different $S$ and $\phi$ in Appendix \ref{app:setting2}).

We compare the proposed fast weight memory programming instruction with normalisation of Sec.~\ref{sec:updaterule}
(denoted here by \textit{ours})
to three baselines: the sum update rule of Sec.~\ref{sec:relation_to_transformers}
(\textit{sum rule}), and two variants of previous update rules  \citep{schlag2020fastweightmemory}: \textit{Schlag (2021)} and \textit{Schlag (2021) with DPFP}.
\textit{Schlag (2021)} is simply the model from \citet{schlag2020fastweightmemory} ported to this setting (i.e. without the LSTM layer).
\textit{Schlag (2021)} has neither a $\phi$ function, nor the sum normalisation term of Sec.~\ref{sec:updaterule}.
Instead it uses a $\tanh$ nonlinearity for its key representations. 
As an ablation we replace it with our DPFP-1 but we don't use the normalisation term of Sec.~\ref{sec:updaterule}, 
which we refer to as \textit{Schlag (2021) with DPFP}.

\begin{figure}[h]
    \begin{center}
        \centerline{\includegraphics[width=\columnwidth]{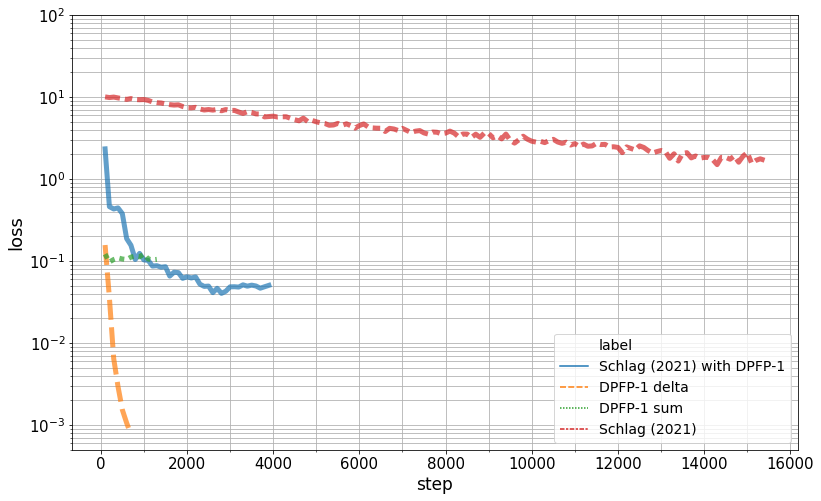}}
        \caption{Learning curves for different update rules. Sequence length of 
        40 and 20 unique keys/values sampled with replacement.}
        \label{fig:setting2}
    \end{center}
    \vskip -0.2in
\end{figure}

Figure \ref{fig:setting2} presents the learning curves.
They demonstrate that our new update rule  outperforms all other variants.
As expected, the baseline sum update rule fails.

\subsection{Machine Translation Experiments}
\label{sec:translation}
Here we compare $\phi$ functions on the standard machine translation task.
We compare Linear Transformer \citep{katharopoulos2020transformers},
Performer \citep{choromanski2020rethinking} and our $\phi$ function DPFP (Sec.~\ref{sec:phi})
to the regular Transformer,  complementing prior comparisons, e.g.,~\citet{tay2020long}.

We use the standard WMT14 English to German Translation dataset and
standard data setups \cite{ott2018, trafo}.
We adapt the recipe of \citet{OttEBFGNGA19} (see Appendix \ref{app:mt})
and train \citet{trafo}'s ``big" models for about 4 days on three V100 GPUs.
We use the exact same training configurations for all
models without model-specific hyper-parameter tuning.
We only vary the model hyper-parameters $m$ in Performers and $\nu$ in DPFP models.

Table \ref{tab:translation} shows the \textsc{Bleu} score \citep{papineni2002bleu, post2018} results.
The Performer is as good as the basic Transformer when the number of samples $m$ is large enough (for $d_\text{dot}=512$, we have $m=256$).
In fact, with $d_\text{key}=64$, the recommended value for $m$ is $d_\text{dot}\log(d_\text{dot})=266$.
Our DPFP model outperforms the Linear Transformer
as well as the Performer when $d_\text{dot}$ is relatively small;
providing a good trade-off between simplicity and performance.

\begin{table}[h]
\caption{WMT14 En-De Translation \textsc{Bleu} scores
for various Transformer models.
Neither model averaging, nor model specific tuning is done.
\textit{Standard} denotes the basic Transformer.}
\label{tab:translation}
\vskip 0.15in
\begin{center}
\begin{small}
\begin{tabular}{lcccccc}
\toprule
       & \multicolumn{3}{c}{Valid}  & \multicolumn{3}{c}{Test} \\  \cmidrule(r){2-4}  \cmidrule(r){5-7}
\multicolumn{1}{c}{$d_{\text{dot}}$} & 64 & 256 & 512 & 64 & 256 & 512 \\
\midrule
Standard   & 26.6 & -     & -    & \textbf{27.7}  & - & - \\
Linear    & 25.5 & -     & -    & 26.8  & - & -\\
Performer & 24.2 & 24.9  & \textbf{26.7} & 24.4  & 25.3 & \textbf{27.7} \\
DPFP (ours)   &  -   & 26.2      &  26.2  &  - & 26.9  & 27.1 \\
\bottomrule
\end{tabular}
\end{small}
\end{center}
\vskip -0.1in
\end{table}

\subsection{Language Modelling Experiments}
\label{sec:lm}
Toy experimental Setting 2 (Sec.~\ref{sec:setting2}) illustrated
the effect of our update rule.
Now our goal is to confirm its effectiveness
on a large-vocabulary word-level language modelling task,
and investigate its further potential.

\paragraph{Experimental setups.}
Our update rule should be evaluated on
a dataset with sufficiently long contextual dependencies.
We use the standard WikiText-103 \citep{merity2016pointer} dataset.
WikiText-103 consists of long articles from Wikipedia;
the training set contains about 28\,K articles with a total of 103\,M running words.
This results in contextual text blocks of about 3600 words.
The validation and test sets also contain similarly long dependencies,
respectively with 218\,K and 246\,K running words for 60 articles each.
The vocabulary size is about 268\,K words.

We split the training data into $L$-word long segments
(which is the backpropagation span).
Unless stated otherwise, we treat these segments independently 
during training.
For evaluation, we use a batch size of one, 
and go through the text with a sliding window of size $L$, taking
into account only the last position for computing perplexity (except
in the first segment where all positions are evaluated). This
is usually done for Transformers with a limited context \cite{al2018character}.
Appendix \ref{app:lm} provides further experimental details.

\begin{table}[t]
\caption{WikiText-103 language model perplexity results
showing effects of our update rule.
The number of trainable parameters are almost the same for all models,
up to the small difference introduced by gating in our update rule (16\,K and 33\,K parameters
respectively for the \textit{small} and \textit{medium} configurations).
We have $D=128$, $L=256$ (40\,M parameters) in the \textit{small},
and $D=256$, $L=384$ (90\,M parameters) in the \textit{medium} configuration.
For Performers, $m$ is $8$ and $16$, respectively.
}
\label{tab:lm}
\vskip 0.15in
\begin{center}
\begin{small}
\begin{tabular}{lcrrrr}
\toprule
     &  Update  & \multicolumn{2}{c}{\textit{small}}  & \multicolumn{2}{c}{\textit{medium}} \\  \cmidrule(r){3-4}  \cmidrule(r){5-6}
          &  Rule  & \multicolumn{1}{r}{Valid} & \multicolumn{1}{r}{Test} & \multicolumn{1}{r}{Valid} & \multicolumn{1}{r}{Test}  \\ \midrule
Transformer   &   -      &  33.0  & 34.1  &  27.9   & 29.6 \\ \midrule
Linear Transformer   &   sum    &  37.1  & 38.3     &  31.1   & 33.0 \\
Delta Network          &  delta    &  \textbf{34.1}  & \textbf{35.5}  & \textbf{29.7} & \textbf{31.5} \\ \midrule
Performer &   sum    &  39.0  & 39.6  & 32.2 & 33.8 \\
          &  delta    &  \textbf{36.1}  & \textbf{37.2}  & \textbf{30.0} & \textbf{31.8} \\
\bottomrule
\end{tabular}
\end{small}
\end{center}
\vskip -0.1in
\end{table}

\paragraph{Effectiveness of our new update rule.}
We first evaluate our update rule in two configurations.
In the \textit{small} configuration,
we set the model dimension (same for key, value, and query) $D$ to 128,
and the training and evaluation context length $L$ to 256.
We note that $D=H * d_\text{dot}$ where $H$ is the number of heads.
$H$ is set to 8. The feed-forward layer dimension is 2048.
The number of layers is 16 in all configurations.
In the \textit{medium} configuration, we set $D=256$ and $L=384$.
Both configurations represent an overcapacity regime.
We evaluate both Linear Transformers \citep{katharopoulos2020transformers}
and Performers \citep{choromanski2020rethinking}.
However, to keep the comparison simple,
we set the capacity of Performers (Sec.~\ref{sec:performer})
equal to the one of linear Transformers,
by the right choice of projection dimension ($m=8$ and $m=16$, respectively,
in \textit{small} and \textit{medium} configurations), even though this limits performance.
We do not include DPFP here, since in both configurations even the smallest value
for $\nu$ provides enough capacity.
Here we investigate the effect of the update rule in an overcapacity scenario
(see Appendix \ref{app:lm_exp} for experimental results in a non-overcapacity regime including DPFP).
All models can be trained using two V100 GPUs in less than four days.
We refer to the Linear Transformer with our delta update rule
as a \textit{Delta Network}.
Table \ref{tab:lm} shows the perplexity results.
In both configurations,
our update rule provides convincing improvements
over the models with the sum update rule.

We also conduct an ablation study to test the effect of
the absolute positional encoding and an extra attention normalisation (Sec.~\ref{sec:updaterule}).
Table \ref{tab:lm-ablation} shows the results.
The \textit{sum normalisation} (Sec.~\ref{sec:updaterule}) is
used in all cases: the models diverged otherwise.
In contrast, better perplexities are obtained when no additional attention normalisation
is applied.
We also observe that the absolute positional encoding is not needed,
confirming results of prior work \citep{irie19:trafolm}.

\begin{table}[h]
\caption{
WikiText-103 language model perplexities
for Linear Transformers (\textit{medium} configuration) with our update rule.
}

\label{tab:lm-ablation}
\vskip 0.15in
\begin{center}
\begin{small}
\begin{tabular}{ccrr}
\toprule
Position Encoding  & Attn. Normalisation  & Valid & Test \\ \midrule
Yes  & Yes   &  30.4   &  32.1   \\ 
\textbf{No}   & Yes   &  29.2   &  31.2   \\ 
Yes  &  \textbf{No}   &  29.7   &  31.5  \\ 
\textbf{No}   &  \textbf{No}   &  \textbf{28.1}   &  \textbf{31.1} \\
\bottomrule
\end{tabular}
\end{small}
\end{center}
\vskip -0.2in

\end{table}

\begin{table}[t]
\caption{
WikiText-103 language model perplexities
when the model is trained and evaluated \textbf{without truncating context},
as opposed to Table \ref{tab:lm} where the context window is limited.
The \textit{medium} config is used.
Neither positional encoding nor attention normalisation
is used for the Delta Net.
The numbers of trainable parameters (Prms.) are given in millions.
We compare with the Transformer-XL at different memory segment lengths.
This results in different state sizes which are proportional to the memory requirements during evaluation,
and highlights the memory efficiency of the Delta Network. 
The state sizes are given in millions.
}
\label{tab:lm-cco}
\vskip 0.15in
\begin{center}
\setlength{\tabcolsep}{0.4em}
\begin{small}
\begin{tabular}{lcccc}
\toprule
Model & Prms.  & State size & \multicolumn{2}{c}{Perplexity}   \vspace{1mm} \\ 
 &  in M. & in M. & Valid & Test   \\ \midrule
Linear Transformer      & 89.8 & 0.13  &  $>$260 $\;$  & $>$260 $\;$ \\
Delta Network  & 89.9 & 0.13 &  27.8   & 29.4  \\  
\midrule
Transformer-XL & 90.9 & 0.13 &  65.7  & 65.5   \\
&  &  1.05 & 29.3  & 30.1  \\ 
& & 2.10  &  26.4  & 27.4  \\  
& & 6.29 &  24.6  & 25.5  \\
\bottomrule
\end{tabular}
\end{small}
\end{center}
\vskip -0.1in

\end{table}

\paragraph{Complexity, wall clock time, memory.}
All methods we propose are within the framework of ``linear Transformers".
Thus, there is no change to be discussed in terms of complexity
which is constant in space and linear in time w.r.t.~sequence length.
However, our modified update rule introduces a few extra computations.
The wall clock time and memory requirement (for the small LM setting)
for the Linear Transformer with and without our delta
update rule are: 63\,K and 66\,K words/sec, and 14 and 13\,GB respectively in our implementation.
The extra resource requirement is thus marginal.
As we use custom CUDA kernels for these linear Transformers,
they are faster than the regular Transformers implemented in PyTorch which process 33K words/sec and require 17\,GB memory.
The speed of the DPFP and Performer models (for Table \ref{tab:lm_non_overcap} in Appendix with a larger $d_\text{dot}$) are 63\,K and 57\,K words/sec.
Performers are slower because of the sampling logic,
which also motivates our DPFP.

\paragraph{Without truncating context.}
Given the constant space requirements,
we can feed inputs to linear Transformers
for an arbitrary number of steps.
To properly assess the 
model's ability to process arbitrary long sequences, it is crucial to
make the training consistent with the evaluation mode \citep{irie19asru}.
During training, we carry over the fast weight memory from one training segment
to the following one, while still limiting the backpropagation span to be within the segment.
We train a Delta Net, using neither positional encoding
nor attention normalisation (the best setting from Table \ref{tab:lm-ablation}).
It was crucial to remove the attention normalisation
for the Delta Net since the accumulator blows up as indicated in Sec.~\ref{sec:updaterule},
while for the Linear Transformer, removing it resulted
in an even worse perplexity of over 1600.
Table \ref{tab:lm-cco} shows the corresponding results.
The Delta Net yields a slight improvement over the best model with a limited context window (Table \ref{tab:lm-ablation}),
unlike the baseline Linear Transformer model with the naive sum update rule which breaks.
We also train a Transformer-XL in our \textit{medium} configuration
as a baseline model specifically designed for this use case
\citep{dai2019transformerxlacl, rae2020compressive}.
We evaluate it using different state sizes by
changing the Transformer XL's memory and target segment lengths (see Appendix \ref{app:lm} for further details).
Performance of the Delta Net does not yet match the
performance of the Transformer XL
when the latter is evaluated with a large state size (large attention window).
However, when we take the state size into account (Table \ref{tab:lm-cco}),
we observe that the Delta Net performs very well
with a small state size,
which is a crucial property in some practical
applications \citep{irie:icassp20}.
These results are promising for future work on alternative
Transformer models which can run for an unlimited number of steps.

\section{Conclusion}
We emphasise the connection between linearised self-attention and Fast Weight Programmers (FWPs, 1991) that program their fast weight memories through sequences of outer products between self-invented key and value patterns. 
The FWP perspective allows for discussing associative memory
capacity limitations of linear attention, and for introducing an alternative differentiable 
elementary programming instruction that the FWP can use to dynamically edit the memory, akin to the famous delta rule, but such that the FWP can learn to use the rule wisely through gradient descent.
We also propose and discuss a new method for linearising  attention.
Experiments on synthetic and real language tasks demonstrate
the effectiveness of our proposals.
The FWP perspective opens up new avenues for investigating even better programming instructions and designs for Transformers with finite memory.

\section*{Acknowledgements}
We thank Sjoerd van Steenkiste, Hubert Ramsauer and Sepp Hochreiter for valuable
comments and suggestions on the first version of the manuscript.
This research was partially funded by ERC Advanced grant no: 742870, project AlgoRNN,
and by Swiss National Science Foundation grant no: 200021\_192356, project NEUSYM.
We thank NVIDIA Corporation for donating several DGX
machines, and IBM for donating a Minsky machine.
We also thank \citet{katharopoulos2020transformers}
for releasing their CUDA implementation of Linear Transformers,
which was helpful to implement our models.

\bibliography{references}
\bibliographystyle{icml2021}

\clearpage
\appendix

\section{Update Rule Derivation}
\subsection{The Update Rule}
\label{app:update_rule}
Here we provide the intermediate steps from Eq.~\ref{eq:updaterule} to Eq.~\ref{eq:updaterule2}.
\begin{align}
\tag{\ref{eq:updaterule}}
\mW^{(i)} &= \mW^{(i-1)}
    \underbrace{+ \vv^{(i)}_\text{new} \otimes \phi(\vk^{(i)})}_{\text{write}} 
    \underbrace{- \bar{\vv}^{(i)} \otimes \phi(\vk^{(i)})}_{\text{remove}} \\
&= \mW^{(i-1)} + \beta^{(i)}(\vv^{(i)} - \bar{\vv}^{(i)}) \otimes \phi(\vk^{(i)})  \tag{\ref{eq:updaterule2}}
\end{align}
By grouping the last two terms, Eq.~\ref{eq:updaterule} becomes:
\begin{align}
\mW^{(i)} &= \mW^{(i-1)} + (\vv^{(i)}_\text{new} - \bar{\vv}^{(i)}) \otimes \phi(\vk^{(i)}) \label{eq:new_minus_bar}
\end{align}

By using the definition of $\vv^{(i)}_\text{new}$ from Eq.~\ref{eq:v_new}:
\begin{align}
\tag{\ref{eq:v_new}}
\vv^{(i)}_\text{new} &= \beta^{(i)} \vv^{(i)} +(1-\beta^{(i)}) \bar{\vv}^{(i)}
\end{align}
we obtain:
\begin{align}
\vv^{(i)}_\text{new} - \bar{\vv}^{(i)} &= \beta^{(i)} \vv^{(i)} +(1-\beta^{(i)}) \bar{\vv}^{(i)} - \bar{\vv}^{(i)} \\
 & =  \beta^{(i)} (\vv^{(i)} - \bar{\vv}^{(i)} )
\end{align}
By substituting this expression to Eq.~\ref{eq:new_minus_bar}, we obtain Eq.~\ref{eq:updaterule2} \qedsymbol.

\subsection{Key Sum Normalisation}
\label{app:norm}

By considering one-hot vectors $\{\ve^{(1)}, ..., \ve^{(i)}, ..., \ve^{(d_\text{key})}\}$
which form the Cartesian basis of $\mathbb{R}^{d_\text{key}}$,
any matrix $\mW \in \mathbb{R}^{d_\text{value} \times d_\text{key}}$ can be written as
\begin{eqnarray}
\mW = \displaystyle \sum_{i=1}^{d_\text{key}} \vw^{(i)}\otimes \ve^{(i)} \label{eq:w_basis}
\end{eqnarray}
where $\{\vw^{(1)}, ..., \vw^{(i)}, ..., \vw^{(d_\text{key})}\}$ are the column vectors of $\mW$.
In the context of associative memory, we can interpret this expression as a set of associations with fixed keys $\ve^{(i)}$ and the associated values $\vw^{(i)}$.

In this view, any update of $\mW$ can be written as updates of each $\vw^{(i)}$.
This perspective allows us to derive the \textit{sum normalisation} of Sec.~\ref{sec:updaterule}.
For that, we start by deriving the update of $\vw^{(i)}$. 

Given an arbitrary weight $\mW$, we consider updating it to $\mW'$ by adding a
new association $(\vk, \vv)$ using our update rule of Sec.~\ref{sec:updaterule} (where we omit
$\beta$):
\begin{eqnarray}
\bar{\vv} &=& \mW \vk \\
\mW' &=& \mW + (\vv - \bar{\vv}) \otimes \vk \label{eq:update_w}
\end{eqnarray}

By substituting $\vk$ in Eq.~\ref{eq:update_w} by its expression in the Cartesian basis $\displaystyle \vk = \sum_{i=1}^{d_\text{key}} k_i \ve^{(i)}$ with $k_i \in \mathbb{R}$, we obtain:
\begin{eqnarray}
\mW' &=& \mW + (\vv - \bar{\vv})  \otimes \sum_{i=1}^{d_\text{key}} k_i \ve^{(i)} \\
&=& \mW + \sum_{i=1}^{d_\text{key}} k_i (\vv - \bar{\vv})  \otimes \ve^{(i)} 
\end{eqnarray}

Now by substituting $\mW$ by its expression of Eq.~\ref{eq:w_basis}:
\begin{eqnarray}
\mW' &=& \sum_{i=1}^{d_\text{key}} \vw^{(i)}\otimes \ve^{(i)}  + \sum_{i=1}^{d_\text{key}} k_i (\vv - \bar{\vv})  \otimes \ve^{(i)}  \\
&=& \sum_{i=1}^{d_\text{key}} \big( \vw^{(i)} +  k_i (\vv - \bar{\vv}) \big) \otimes \ve^{(i)}
\end{eqnarray}
The column-wise update is thus: 
\begin{eqnarray}
\vw'^{(i)} = \vw^{(i)} +  k_i (\vv - \bar{\vv}) \label{eq:col_update1}
\end{eqnarray}

We can explicitly write down $\bar{\vv}$ as:
\begin{eqnarray}
\bar{\vv} = \mW \vk = \mW \sum_{j=1}^{d_\text{key}} k_j \ve^{(j)} = \sum_{j=1}^{d_\text{key}} k_j \vw^{(j)}
\end{eqnarray}
which we can substitute in Eq.~\ref{eq:col_update1} to obtain:
\begin{eqnarray}
\vw'^{(i)} &=& \vw^{(i)} +  k_i (\vv - \sum_{j=1}^{d_\text{key}} k_j \vw^{(j)} ) \\
&=& \vw^{(i)} +  k_i \vv - \sum_{j=1}^{d_\text{key}} k_i k_j \vw^{(j)} \label{eq:col_update_final}
\end{eqnarray}
In Eq.~\ref{eq:col_update_final}, the \textit{weight} $k_i$ on the positive term $\vv$ is in general
not equal to the total weights on the negative terms $\sum_{j=1}^{d_\text{key}} k_i k_j$.
We can force these weights to be balanced by introducing the normalisation: $ \displaystyle \sum_{j=1}^{d_\text{key}} k_i k_j = k_i$.

If $k_i$ is non zero, we obtain: 
\begin{eqnarray*}
\sum_{j=1}^{d_\text{key}} k_j  = 1
\end{eqnarray*}
This corresponds to the \textit{sum normalisation} we introduced in Sec.~\ref{sec:updaterule} \qedsymbol.

\section{Formal comparison to \citet{peng2021random}}
\label{sec:peng}
Concurrently to our work, \citet{peng2021random} proposed the following gated update rule:
\begin{align}
\mW^{(i)}= (1-\beta^{(i)}) \mW^{(i-1)} + \beta^{(i)} \vv^{(i)} \otimes \phi(\vk^{(i)})
\end{align}
which is motivated by the gating mechanism in recurrent neural networks \citep{hochreiter1997long}.
In contrast, our update rule of Eq.~\ref{eq:updaterule2} 
\begin{align}
\mW^{(i)}= \mW^{(i-1)} + \beta^{(i)}(\vv^{(i)} - \bar{\vv}^{(i)}) \otimes \phi(\vk^{(i)})  \tag{\ref{eq:updaterule2}}
\end{align}
is driven by an associative memory
perspective, relates to the famous error-correcting delta rule,
and offers a crucial property.

To illustrate a similarity and a crucial difference between the two update rules,
we consider a fast weight matrix $\mW$ which is constructed by two associations $(\vk_1, \vv_1)$
and $(\vk_2, \vv_2)$, i.e.
\begin{align}
\mW = \vv_1 \otimes \vk_1 + \vv_2 \otimes \vk_2
\end{align}
where we assume $\vk_1$ and $\vk_2$ to be orthonormal, and we omit $\phi$.
Now we consider updating $\mW$ to $\mW'$ by adding a new association $(\vk_3, \vv_3)$ where $\vk_3=\vk_2$.
Using \citet{peng2021random}'s update rule, we have:
\begin{align*}
\mW'= (1-\beta) \mW + \beta \vv_3 \otimes \vk_3 
\end{align*}
This rule thus updates the value associated with the key $\vk_2=\vk_3$ to be a convex combination of the old and the new values $(1-\beta) \vv_2 + \beta \vv_3$:
\begin{align*}
\mW' \vk_3 &= (1-\beta) \mW \vk_3 + \beta \vv_3 \\
&= (1-\beta) \vv_2 + \beta \vv_3 
\end{align*}
However, it also modifies or in the worst case erases the value associated with the key $\vk_1$:
\begin{align*}
\mW' \vk_1 &= (1-\beta)  \mW \vk_1 = (1-\beta)  \vv_1
\end{align*}

In contrast, using our update rule, we have:
\begin{align*}
\mW'= \mW + \beta (\vv_3 - \vv_2) \otimes \vk_3 
\end{align*}
since $\bar{\vv} = \mW \vk_3 = \mW \vk_2 = \vv_2$.\\
Our rule thus also updates the value associated with the key $\vk_2=\vk_3$ to be a convex combination of the old and the new values $(1-\beta) \vv_2 + \beta \vv_3$:
\begin{align*}
\mW' \vk_3 &= \mW \vk_3 + \beta (\vv_3 - \vv_2) \\
& = \vv_2 + \beta (\vv_3 - \vv_2) \\
&=(1-\beta) \vv_2 + \beta \vv_3
\end{align*}
while crucially, it keeps the value associated with $\vk_1$ unmodified:
\begin{align*}
\mW' \vk_1 &= \mW \vk_1 = \vv_1
\end{align*}
Our update rule thus differs from \citet{peng2021random}'s one on this property of updating associations while
keeping other ``unrelated" ones intact in an associative memory.

\section{DPFP-$\nu$ Implementation}
\label{app:dpfp}

Listing \ref{lst:dpfp} is a simple PyTorch implementation of DPFP-$\nu$ (Eq.~\ref{eq:dpfp}) which consist of two concatenations followed by one element-wise multiplication.

\begin{lstlisting}[
language=python, 
caption={Simple PyTorch implementation of DPFP-$\nu$ (Eq.~\ref{eq:dpfp}).},
label={lst:dpfp}]
import torch
from torch import cat
from torch.nn.functional import relu as r

def dpfp(x, nu=1):
  x = cat([r(x), r(-x)], dim=-1)
  x_rolled = cat([x.roll(shifts=j, dims=-1)
           for j in range(1,nu+1)], dim=-1)
  x_repeat = cat([x] * nu, dim=-1)
  return x_repeat * x_rolled
\end{lstlisting}

 \section{Additional Experimental Results}
In this section, we provide additional experimental results which
we could not include in the main paper because of space limitations.

\subsection{Synthetic Task Setting 1}
\label{app:setting1}
\begin{figure}[h]
    \begin{center}
        \vskip -0.1cm
        \centerline{\includegraphics[width=\columnwidth]{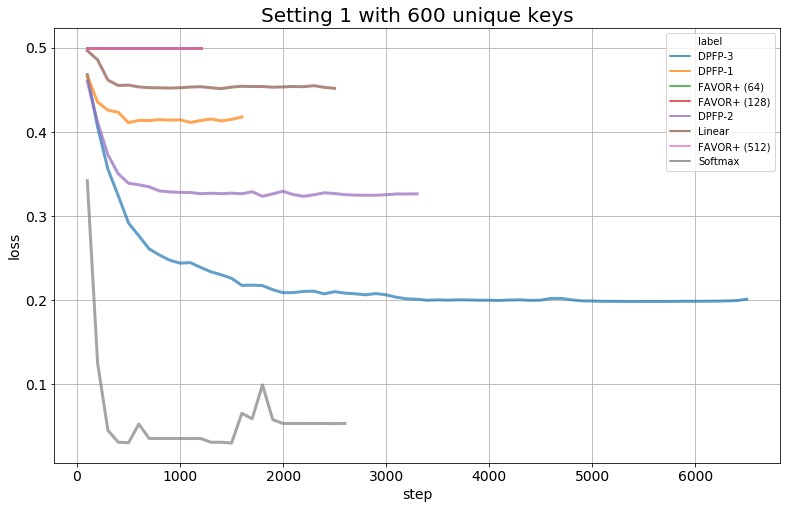}}
        \vskip -0.1cm
        \caption{Training curves for setting 1 with 600 unique keys/values (sampled without replacement) as described in Sec.~\ref{sec:setting1}.}
        \label{fig:setting1_600}
    \end{center}
    \vskip -0.5cm
\end{figure}

Figure \ref{fig:setting1_600} shows learning curves for the synthetic setting 1 (without replacement) with 600 unique keys and values.
The scripts used to generate such figures can be found in our GitHub repository.

\subsection{Synthetic Task Setting 2}
\label{app:setting2}
\begin{figure}[h]
    \begin{center}
        \vskip -0.1cm
        \centerline{\includegraphics[width=\columnwidth]{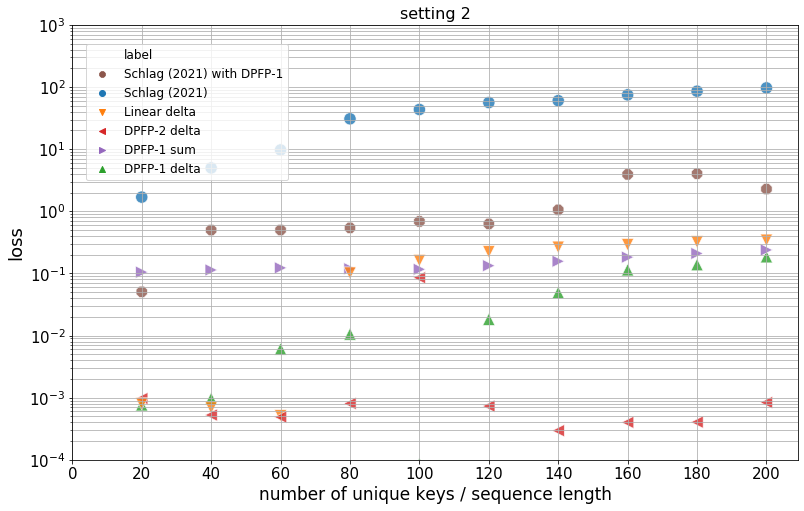}}
        \vskip -0.1cm
        \caption{Final evaluation loss on synthetic setting 2 (with replacement) problems with the total number of unique associations ranging from 20 to 200. Each individual symbol is a model trained until convergence as described in Sec.~\ref{sec:setting2}. In all problems, with different sequence lengths and a different number of unique keys, our update rule outperforms all other approaches.}
        \label{fig:setting2_all}
    \end{center}
    \vskip -0.5cm
\end{figure}

Figure \ref{fig:setting2_all} is a capacity plot for setting 2 with an increasing number of unique keys and queries (analogous to Figure \ref{fig:setting1} of setting 1 apart from the log-scale of the y-axis).
We did not include FAVOR+ in this plot, because its combination with our update rule resulted in not-a-number
in this setting.

\subsection{Language Modelling}
\label{app:lm_exp}
In Sec.~\ref{sec:lm}, we evaluated our update rule when the model is under overcapacity regime.
Here we present an extra language modelling experiment
which evaluate the benefits of our update rule in non-overcapacity scenarios.
This also allows us to include DPFP in the evaluation.
We train both, Performer and DPFP, in the \textit{small} setting ($D=128$, $L=256$)
with $m=16$ and $\nu=1$, resulting in $d_\text{dot} = 256$ for both cases.
Table \ref{tab:lm_non_overcap} shows the perplexity results.
First we observe that the Performer and DPFP baseline models with the sum update rule
do not outperform the Linear Transformer baseline from Table \ref{tab:lm}.
In fact, language modelling might be less affected by the capacity issue
than the synthetic retrieval task, as it might not require the exact retrieval.
Second we observe that our update rule improves both variants of linear attention
over the sum update-rule baselines even in this condition.
This indicates the general benefits of our update rule in Fast Weight Programmers.
We note that the improvement is larger for the DPFP model than
for the Performer.
This is similar to Table \ref{tab:lm} where our update rule
improves the deterministic Linear Transformers more than the Performers.
Finally, we note that we also tried the DPFP and Performer models
with an increased $d_\text{dot}$ by setting $\nu=2$
and $m=32$ respectively. 
While this increases $d_\text{dot}$ by a factor of two,
it was not beneficial for this language modelling setting.

\begin{table}[h]
\caption{WikiText-103 language model perplexity results
showing effects of our update rule
in non-overcapacity regime.
The number of trainable parameters are almost the same for all models,
up to the small difference introduced by gating in our update rule (16\,K parameters).
The \textit{small} config is used, i.e. $D=128$, $L=256$ (40\,M parameters).
We set $m=16$ for the Performers and $\nu=1$ for the DPFP models,
which result in $d_\text{dot} = 256$ for both cases.
The model is thus not necessary in an overcapacity regime.  
}
\label{tab:lm_non_overcap}
\vskip 0.15in
\begin{center}
\begin{small}
\begin{tabular}{l|c|rr}
\toprule
     &  Update  & \multicolumn{2}{c}{\textit{small}}  \\
          &  Rule  & \multicolumn{1}{r}{Valid} & \multicolumn{1}{r}{Test}  \\ \midrule
Transformer   &   -      &  33.0  &  34.1 \\ \midrule
Performer &   sum    & 38.0   & 38.8 \\
          &  delta    & \textbf{36.0}  &  \textbf{37.0} \\ \midrule
DPFP &   sum    &  37.7  & 38.8   \\
          &  delta    &  \textbf{33.9}  & \textbf{35.0}  \\
\bottomrule
\end{tabular}
\end{small}
\end{center}
\vskip -0.1in

\end{table}

\section{Details on Machine Translation Experiments}
\label{app:mt}
We implemented different $\phi$ functions in the \textsc{fairseq} tookit \cite{OttEBFGNGA19}.
The Transformer architecture used in the experiment
is the one referred to as \textit{big} in the original Transformer paper \cite{trafo}:
the model has 6 layers each in the encoder and the decoder, with a hidden layer size of 1024
with 16 attention heads, 4096-dimensional feed-forward layers,
using 32\,K byte-pair encoding sub-word units \cite{sennrich16bpe}.
\textsc{fairseq} provides a training configuration for the corresponding model \cite{ott2018},
which we adapted for our infrastructure.
We trained our models on three GPUs using a batch size of up to 3584 tokens per GPU
and accumulating gradients over 16 batches for 45 epochs, and selected the best model
based on the validation \textsc{Bleu} score.
In Table \ref{tab:translation}, we directly report \textsc{Bleu} for different values of $d_\text{dot}$;
Table \ref{tab:dot-dim} provides the conversion from hyper-parameters $m$ of Performers or $\nu$ in the DPFP to $d_\text{dot}$.

\begin{table}[h]
\caption{Relation between dot product space dimension
and the hyper-parameters in the Performer and our DPFP models.
$d_{\text{key}}=64$ in all our translation models.}
\label{tab:dot-dim}
\vskip 0.15in
\begin{center}
\begin{small}
\begin{tabular}{lcccr}
\toprule
$d_{\text{dot}}$ & 256 & 384 & 512 \\
\midrule
Performer $m$ & 128 & 192 & 256  \\  
DPFP $\nu$    & 2 & 3 & 4 \\ 
\bottomrule
\end{tabular}
\end{small}
\end{center}
\vskip -0.1in
\end{table}

\section{Details on Language Modelling Experiments}
\label{app:lm}
\paragraph{Implementation notes.}
All our implementations are based on PyTorch \citep{NEURIPS2019_bdbca288}.
Our base language modelling code has been developed by using the public code by \citet{dai2019transformerxlacl} for Transformer-XL as a starting point.
For $\phi$ functions, we ported the same implementation we used for our
translation experiments.
For the implementation of our update rule,
we modified the CUDA kernel for the Linear Transformer made publicly available by \citet{katharopoulos2020transformers}.
We note that a custom implementation of the backward pass for fast weights
is crucial for language modelling.
A naive backward computation generated by automatic differentiation would store the fast weights for each time step, which can quickly hit the GPU memory limit.
The custom implementation ensures that we need to store
only one set of weights by recomputing the fast weights needed for computing
the gradients for each time step in the backward pass (which still remains time-efficient as the operations involved in the computation of our fast weights are rather inexpensive). 

\paragraph{Experimental details.}
Here we provide extra experimental details to complement
the descriptions of Sec.~\ref{sec:lm}.
For the \textit{small} and \textit{medium} configurations,
we use batch sizes of 96 and 56 sequences, respectively,
and train for about 120 and 70 epochs.
In both settings, we apply 10\% dropout \citep{hanson1990stochastic, srivastava2014dropout},
and train using the Adam optimiser \citep{kingma2014adam} with an initial learning rate of 0.00025 and 2000 learning rate warm-up steps.
For further details, we refer the readers to our code.
For experiments with Transformer-XL (Table \ref{tab:lm-cco}),
we train it with the same backpropagation span as our models
(i.e. $384$ words in the medium configuration).
The model is trained with memory and target segment lengths
of 384.
The models with different state sizes in Table \ref{tab:lm-cco} are obtained by using different Transformer-XL memory segment lengths at evaluation time.
The models with state sizes of 1.05~M, 2.10~M, and 6.29~M are obtained
by using memory and target lengths of 64, 128, and 384, respectively.
The model with a state size of 0.13~M uses a memory length of 15 and a target length of 1.
Like for other models, a batch size of 1 is used
for evaluating the Transformer XL.
The state sizes in Table \ref{tab:lm-cco} are computed
as follows.
The per-layer state size of the Linear Transformer and the Delta Net
are: number of heads (here 8) $\times$ fast weight matrix size which is per-head key dimension
(here 32) $\times$ per-head value dimension (here 32).
This yields a total size of 8,192.
The per-layer state size of the Transformer XL is:
memory segment length $\times$ target segment length 
$\times$ (total key dimension, here 256 $+$ total value dimension, here 256).
We obtain the total state size we report in Table \ref{tab:lm-cco} by multiplying the 
per-layer state size by the number of layers which is 16 for all our models.




\end{document}